\documentclass[letterpaper]{article} 
\usepackage{aaai25}  
\usepackage{times}  
\usepackage{helvet}  
\usepackage{courier}  
\usepackage[hyphens]{url}  
\usepackage{graphicx} 
\usepackage{natbib}  
\usepackage{caption} 
\usepackage{algorithm}
\usepackage{algorithmic}

\usepackage{newfloat}
\usepackage{listings}
\DeclareCaptionStyle{ruled}{labelfont=normalfont,labelsep=colon,strut=off} 
\floatstyle{ruled}
\newfloat{listing}{tb}{lst}{}
\floatname{listing}{Listing}
\usepackage{bbding}
\usepackage{amsmath}
\usepackage{pifont}
\usepackage{amssymb}
\usepackage{multirow}
\makeatletter

\title{Temporal Action Localization with Cross Layer Task Decoupling and Refinement}
\author{
Qiang Li\textsuperscript{\rm 1,\rm 4}\equalcontrib,
Di Liu\textsuperscript{\rm 1,\rm 2}\equalcontrib,
Jun Kong\textsuperscript{\rm 1,\rm 3}\footnotemark[2],
Sen Li\textsuperscript{\rm 1},
Hui Xu\textsuperscript{\rm 4},
Jianzhong Wang\textsuperscript{\rm 1}\thanks{Corresponding author.}
}

\affiliations{
\textsuperscript{\rm 1}Northeast Normal University,             
\textsuperscript{\rm 2}Northeast Electric Power University\\  
\textsuperscript{\rm 3}KLAS of MOE,       
\textsuperscript{\rm 4}Changchun Humanities and Sciences College\\

\{ liq782, kongjun, lis084, wangjz019 \}@nenu.edu.cn, 20102313@neepu.edu.cn, xuhui1@ccrw.edu.cn
}

\usepackage{bibentry}

\begin{document}
\nocopyright
\maketitle

\begin{abstract}
	Temporal action localization (TAL) involves dual tasks to classify and localize actions within untrimmed videos. However, the two tasks often have conflicting requirements for features. Existing methods typically employ separate heads for classification and localization tasks but share the same input feature, leading to suboptimal performance. To address this issue, we propose a novel TAL method with Cross Layer Task Decoupling and Refinement (CLTDR). Based on the feature pyramid of video, CLTDR strategy integrates semantically strong features from higher pyramid layers and detailed boundary-aware boundary features from lower pyramid layers to effectively disentangle the action classification and localization tasks. Moreover, the multiple features from cross layers are also employed to refine and align the disentangled classification and regression results. At last, a lightweight Gated Multi-Granularity (GMG) module is proposed to comprehensively extract and aggregate video features at instant, local, and global temporal granularities. Benefiting from the CLTDR and GMG modules, our method achieves state-of-the-art performance on five challenging benchmarks: THUMOS14, MultiTHUMOS, EPIC-KITCHENS-100, ActivityNet-1.3, and HACS.
	Our code and pre-trained models are publicly available at: {https://github.com/LiQiang0307/CLTDR-GMG}.
\end{abstract}

%

\section{Introduction}

With the aim of accurately identifying action categories and localizing its start and end times within a video, temporal action localization (TAL) has garnered significant attention from the research community due to its wide applications, including intelligent surveillance, video summarization, highlight detection, and visual question answering.

Recent advancements in deep learning techniques have led to remarkable progress in various computer vision fields, including TAL. Numerous deep learning methods have been proposed to address the TAL problem, such as~\cite{S-CNN,SS-TAD,GTAN}. A typical deep learning-based TAL model comprises three main steps: extracting video features using deep learning networks, encoding these features into a multi-layer feature pyramid to capture action information at different temporal scales and using a decoder with classification and regression heads for action recognition and localization. The initial step of video feature extraction is often accomplished using pretrained video classification backbones. Consequently, the encoder and decoder in the subsequent steps become critical for achieving high TAL performance. To model the temporal context of features, various techniques have been employed in TAL, including 1D CNN \cite{AFSD}, GCN~\cite{G-TAD}, xGPN \cite{20} and RNN \cite{21}. Recently, the Transformer \cite{Transformer} with self-attention mechanism and its variants have also been adopted to capture long-range temporal dependency between video features~\cite{ActionFormer,TemporalMaxer,Tridet}. Nevertheless, the existing TAL methods still encounter two limitations.
\begin{figure}[]
	\centering
	\includegraphics[width=1.0\columnwidth]{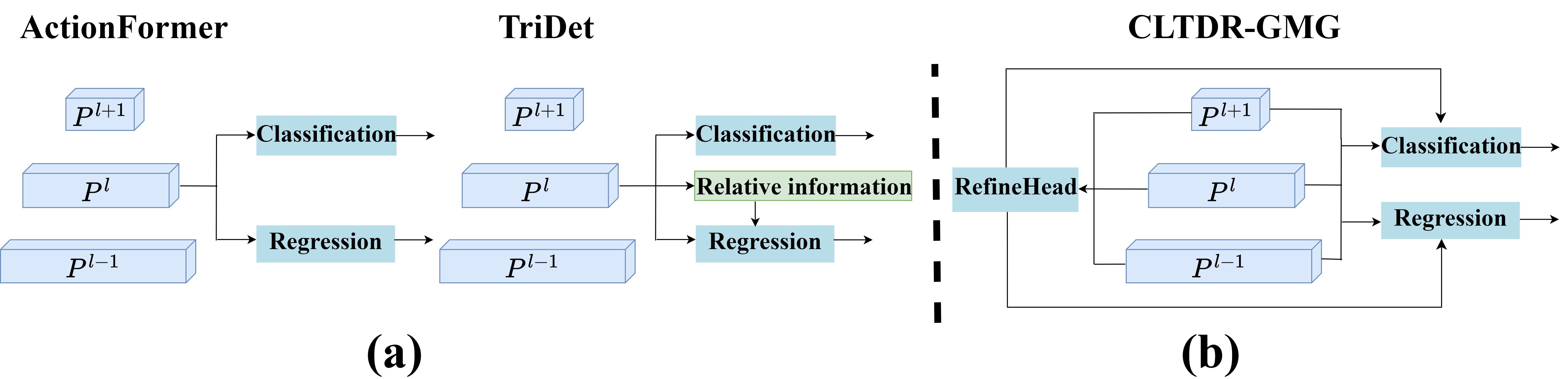} 
	\caption{Comparison of different task decoupling. (a) Previous methods use two  head branches share the same input feature; (b) Our method uses cross layer features for task decoupling and refinement. Zoom in for better view.}
	\label{fig1}
\end{figure}

On the one hand, current methods often struggle to balance the trade-off between classification and localization. Typically, these methods adopt two separated convolutional heads at each layer of video feature pyramid to accomplish the category classification and boundary regression of actions. However, this strategy relies heavily on learned parameters in the convolutional networks to decouple tasks because both heads process identical features at each pyramid layer~\cite{ActionFormer}. Although TriDet~\cite{Tridet} incorporates relative information to help boundary prediction, this information is still extracted independently at each feature pyramid layer (Fig. 1a). Thus, its overall performance may still be suboptimal. Moreover, existing TAL methods often lack explicit interaction between classification and localization heads, which would exacerbate the inconsistency between their predictions. 

On the other hand, the features encoded by existing methods are incomprehensive since most of them only consider the local temporal information of the video. Though the global temporal context can be captured by self-attention, it is achieved at the cost of high computational overhead. Therefore, a local self-attention strategy is adopted by some studies to restrict the attention within a limited temporal window~\cite{ActionFormer, Tallformer}. While both the local and instant temporal information is aggregated in TriDet~\cite{Tridet}, the global temporal dependencies among the entire video are still neglected.

To alleviate the above limitations and improve the TAL performance, a novel method is presented in this paper. 

First, we introduce a Cross-Layer Task Decoupling and Refinement (CLTDR) strategy to disentangle classification and localization tasks and enhance their consistency. Specifically, the CLTDR based decoder incorporates suitable cross layer features to handle the specific tasks in TAL. As shown in Fig. 1b, since the action classification requires rich semantic context information for accurate category inference, we combine the feature at each pyramid layer with temporally-coarse yet semantically-strong feature from higher layer for this task. Conversely, the action localization necessitates highly detailed information for boundary regression. Hence, we fuse the feature of each pyramid layer with a finer feature extracted from lower layer to address this task. During the two cross layer feature fusion, the attention mechanism is utilized to select important information from higher and lower layer features. Furthermore, we also incorporate a refinement head that leverages both higher and lower layer features to harmonize the decoupled classification and localization tasks, thereby the consistency and accuracy of their respective outputs can be enhanced.

Second, we propose a Gated Multi-Granularity (GMG) module at each feature pyramid layer to comprehensively aggregate the information from multiple temporal granularities. The major characteristic of GMG module is that it utilizes three-branch network as a substitution for self-attention in Transformer architecture. In this network, we adopt simple fully-connected operation and 1D depth-wise convolution to obtain the instant and local temporal information. Additionally, a 1D Fast Fourier Transform (FFT) branch with a learnable global filter is employed to capture global temporal dependencies across the video in frequency domain, offering a more efficient alternative to self-attention. To mitigate potential redundancy among features extracted from different temporal granularities, we employ a gated mechanism to selectively fuse features of various branches and improve the discriminative ability of the final representation.

Considering the aforementioned two major contributions, the proposed method is termed CLTDR-GMG. Numerous experiments on five datasets demonstrate that our method achieves state-of-the-art TAL performance.

\section{Related Work}
\begin{figure*}[htpb]
	\centering
	\includegraphics[width=1\textwidth]{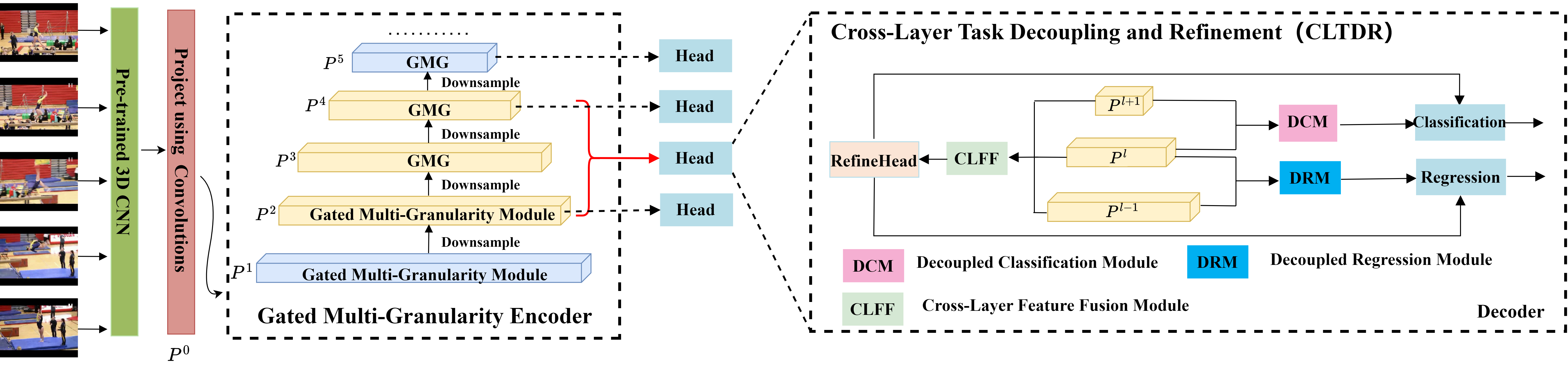} 
	\caption{An illustration of our method. We build a feature pyramid with GMG module. The CLTDR decoder at the \textit{l}-th pyramid layer leverages the features $P^{l+1}$ and $P^{l-1}$ to generate distinct representations for classification and localization tasks, followed by a refinement using RefineHead.}
	\label{fig2}
\end{figure*}
\begin{figure}[htpb]
	\centering
	\includegraphics[width=1.0\columnwidth]{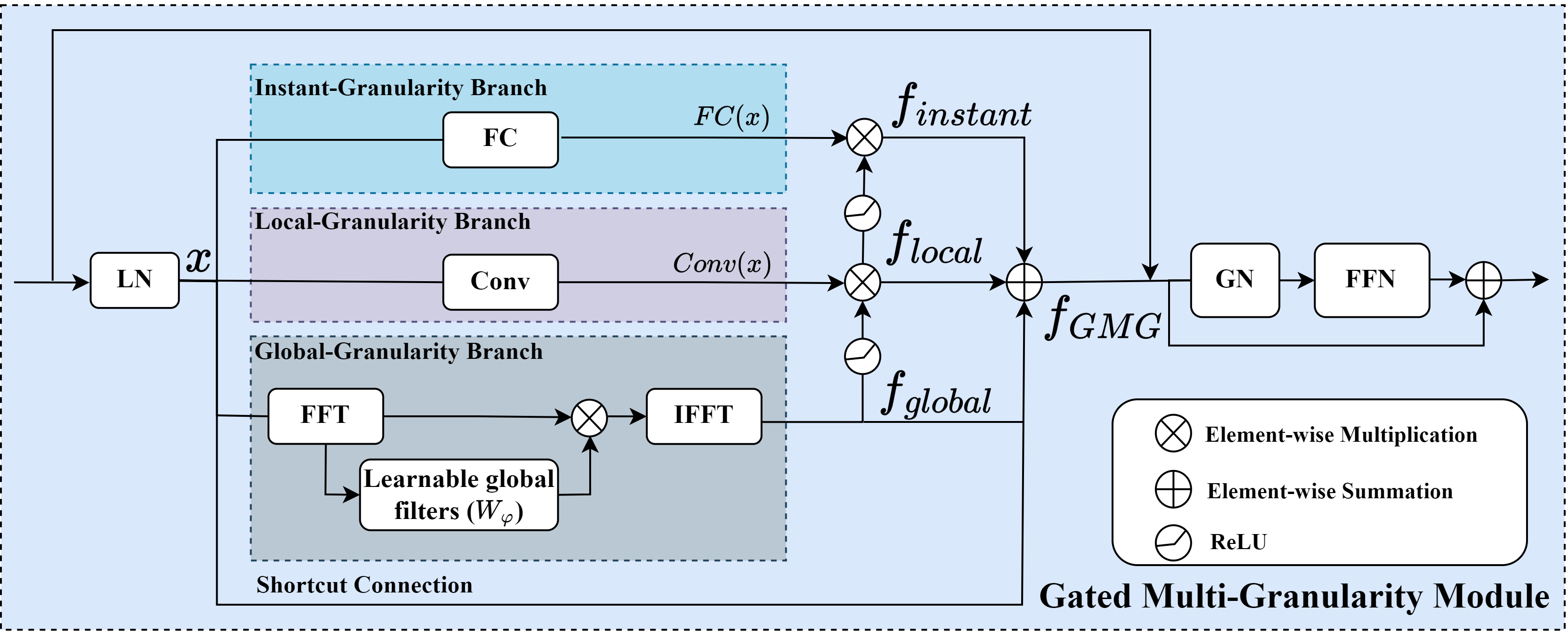} 
	\caption{ Illustration of GMG. Zoom in for better view.}
	\label{fig3}
\end{figure}
\textbf{Temporal Action Localization (TAL).} The existing TAL methods primarily fall into two categories: two-stage methods and one-stage methods. Two-stage TAL initially generates proposals to localize potential action instances before classifying them into different categories. Therefore, most two-stage methods prioritize proposal generation. Various techniques are employed for this purpose, including anchor window classification \cite{36}, action boundaries detection \cite{BSN}, graph based representation learning \cite{BC-GNN}, and fine-grained temporal representation design ~\cite{41}. Despite their effectiveness, two-stage methods face significant challenges, including high computational complexity and the inability to be end-to-end optimized. One-stage TAL methods combine action localization and classification into a single framework by eliminating the proposal generation stage. Some previous studies ~\cite{AFSD,14,SSAD} accomplished the one-stage TAL by constructing a hierarchical network architecture with CNNs. Subsequently, anchor-free model \cite{GTAN} has been introduced as a simplified approach for one-stage TAL. More recently, some innovative techniques, such as Transformer \cite{Transformer}, SGP \cite{Tridet} and Mamba \cite{mamba} have also been leveraged to further improve the one-stage TAL performance. However, as mentioned in the Introduction section, there still exist certain limitations in these methods.

\textbf{Task Decoupling.} The conflict between classification and regression tasks, initially observed in object detection, has led to a widespread adoption of decoupled heads in object detectors. Double head RCNN \cite{RCNN} introduced separate heads for classification and localization within the RCNN framework. YOLOX \cite{yolox} adopted decoupled heads to address the detrimental impact of coupled classification and localization tasks for YOLO series algorithms. DDOD \cite{DODD} proposed an adaptive disentanglement module that employs deformable convolution to automatically obtain beneficial features for classification and regression. These studies underscore the importance of decoupling classification and localization tasks. However, the task decoupling in them occurs solely at the parameter level with the same input features, leading to an imperfect trade-off between the two tasks. To address this issue, ~\cite{tscode} proposed the task-specific context decoupling (TSCODE) head, which further disentangles the classification and localization tasks at feature level. 

Similar to object detection, existing one-stage TAL methods employ separate heads for action classification and localization with shared input feature. Thus, the task decoupling in them is still limited to the parameter level. Additionally, the classification and localization heads in existing methods work independently, which may lead their results to be inconsistent and inaccuracy.
\section{Method}
\subsection{Problem Statement}
We assume that an untrimmed video with $T$ frames can be represented by a set of feature vectors $X = \left\{ x_{1},\cdots,x_{T} \right\} $, where $x_{t}$ is the feature of frame $t$. Temporal action localization aims at detecting a set of action instances $Y = \left\{ y_{1},\cdots,y_{N} \right\}$ from $X$, where $N$ is the number of actions. The $n$-th action in $Y$ is defined as $y_{n} = \left( s_{n},e_{n},a_{n} \right)$, where $s_{n} \in \left\lbrack {1,T} \right\rbrack$, $e_{n} \in \left\lbrack {1,T} \right\rbrack$ and $a_{n} \in \left\{ {1,\ldots,C} \right\}$ denote its start time, end time and corresponding label, respectively. $C$ is the number of action categories.

\subsection{Method Overview}
As depicted in Figure 2, our CLTDR-GMG architecture comprises three key components: First, a pre-trained 3D CNN network is employed to extract spatiotemporal features from the input video. Subsequently, an encoder formed by the proposed GMG module and downsampling is utilized to construct a feature pyramid that captures actions across various temporal scales. Finally, a CLTDR module decodes the feature within each pyramid layer for action localization. A detailed explanation of our core innovations, i.e., GMG and CLTDR, will be elaborated in the following section.
\begin{figure*}[t]
	\centering
	\includegraphics[width=0.9\textwidth]{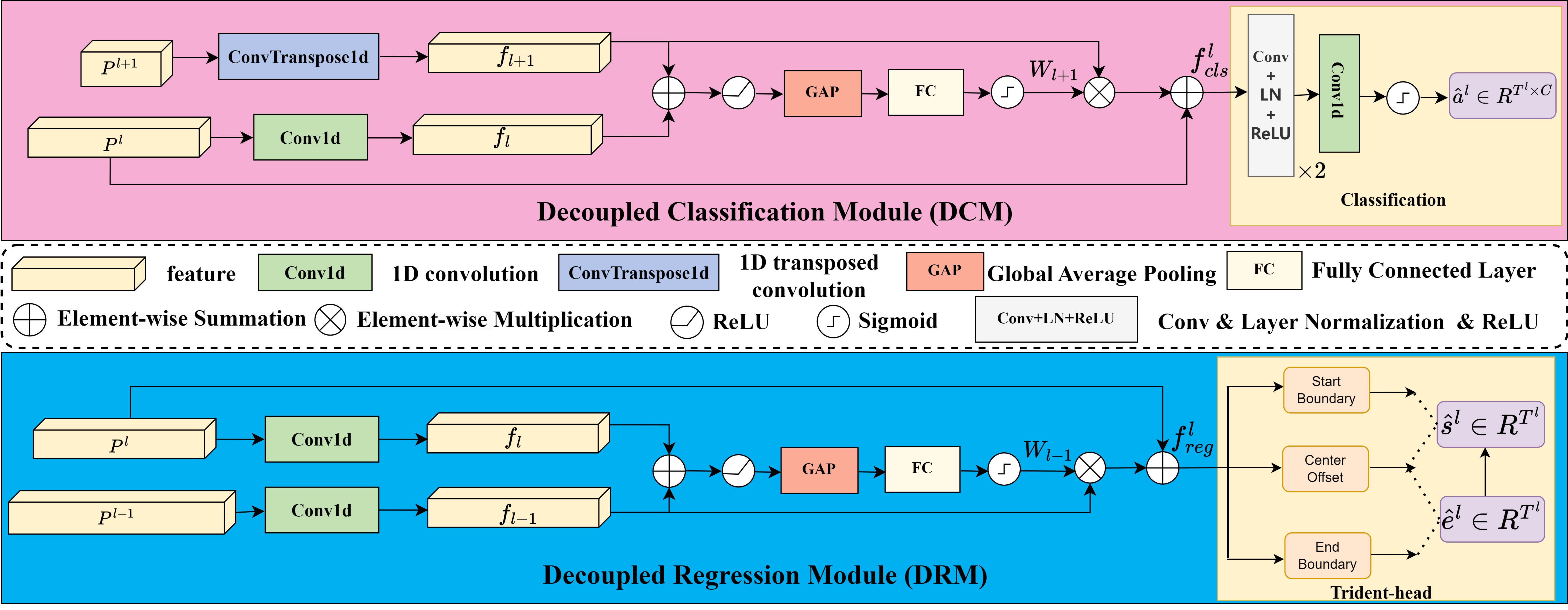} 
	\caption{Illustration of Decoupled Classification Module and Decoupled Regression Module.}
	\label{fig4}
\end{figure*}
\subsection{Encoder with GMG}
Our approach begins by encoding the video feature $X$ into a multi-scale feature pyramid $P = \left\{ P^{1},P^{2},\cdots P^{L} \right\}$. The encoder comprises two primary components: a simple 1D CNN for initially feature projection and our proposed GMG module for aggregating information across diverse temporal granularities. 

To facilitate local context learning within time series data and promote stable training \cite{ActionFormer}, we employ two consecutive 1D CNNs followed by ReLU activation to embed  video feature $X$ into a $D$ dimensional space. The projected feature of $X$ is denoted as $P^{0} \in \mathbb{R}^{D \times T}$.

Following feature projection, a feature pyramid is constructed using the proposed Gated Multi-Granularity(GMG) module. Briefly, the initial feature $P^{0}$ is iteratively downsampled to a series of smaller temporal scales and the feature in each scale are then processed by GMG module to capture instant, local and global temporal information. As shown in Figure 3, GMG utilizes a simple fully-connected ($FC$) and a 1D depth-wise convolution ($Conv$) network for instant and local feature extraction, respectively. To exploit global feature dependencies, we employ Fast Fourier Transform (FFT), a powerful tool for global feature extraction, as validated by recent research~\cite{FNet,GFNet,AFNO}. 
Given the input sequence $x$ with $T$ time steps, we first transform it to the frequency representations by FFT for global information aggregation:
\begin{equation}
	X_{f}\lbrack u\rbrack = FFT(x) = {\sum\limits_{t = 0}^{T - 1}{x_{t}e^{- j\frac{2\pi}{N}ut}}},0 < u < T - 1
\end{equation}

Subsequently a global filter $W_{\varphi}$  whose dimension is equivalent to $X_{f}$ is learned by a convolution, followed by a ReLU activation and another convolution as 
\begin{equation}
	W_{\varphi} = Conv\left( ReLU\left( Conv\left( X_{f} \right) \right) \right)
\end{equation}
where $Conv$ and $ReLU$ denote the 1D depth-wise convolution and activation function.

Finally, we get the updated feature representations in the original space by
\begin{equation}
	f_{global} = IFFT\left( X_{f} \otimes W_{\varphi} \right)
\end{equation}
where $\otimes$ denotes element-wise multiplication (i.e., Hadamard product) and IFFT is inverse FFT. Based on the equivalence between Hadamard product in Fourier domain and convolution in time domain~\cite{oppenheim1999discrete}, the feature $f_{global}$ could be interpreted as the outputs of a convolution whose kernel dimension matches the number of feature vectors in $x$.

Since the feature extracted from larger temporal granularity may inherently contain some information in smaller granularities, we incorporate a gated mechanism with $ReLU$ activation to selectively emphasize discriminative features and suppress redundant information within the instant and local branches. As a result, the feature of GMG can be obtained by combining the original feature $x$ with the features of the three granularity branches as

\begin{scriptsize}
	\begin{equation}
		f_{GMG} = x + \underbrace{ReLU(f_{local}) \otimes FC(x) }_{f_{instant}}+ \underbrace{ReLU(f_{global}) \otimes Conv(x)}_{f_{local}} + f_{global}
	\end{equation}
\end{scriptsize}

After multi-granularity feature extraction, our GMG module incorporates a Group Normalization (GN) and a Feed Forward Network (FFN) with residual connections to further process the feature $f_{GMG}$. To construct the feature pyramid, 1D max-pooling is employed for downsampling between adjacent layers.

\subsection{Decoder with CLTDR}

The GMG based encoder generates a feature pyramid $P = \left\{ P^{1},P^{2},\cdots P^{L} \right\}$. To decode this feature pyramid into the sequence label $\hat{Y} = \left\{ {\hat{y}}_{1},{\hat{y}}_{2},\cdots,{\hat{y}}_{N} \right\}$, we propose a Cross-Layer Task Decoupling and Refinement (CLTDR) module that disentangles features for classification and localization tasks. The CLTDR consists of three main components: a decoupled classification module, a decoupled regression module, and a Refinehead. 

\textbf{Decoupled Classification Module.} Some studies have demonstrated the sufficiency of sparsely distributed key frames for action classification~\cite{yan2018deep,dong2022identifying}. However, existing TAL methods often neglect the redundant and uninformative feature in the classification task. Moreover, the classification accuracy of individual frame benefits from context information from its surrounding frames. For example, the athletic actions like Long Jump and High Jump have the same initial running. Thus, it is hard to identify the categories of the two actions without the information in the following frames. These observations highlight the importance of temporally coarse yet semantically robust features for classification task.

Based on above analysis, we propose a decoupled classification module (DCM) as illustrated in Figure 4. Specifically, the feature at each pyramid layer is combined with its coarse abstraction, i.e., the feature extracted from higher layer, to generate a semantically rich feature for classification. Taking feature $P^{l} \in R^{{D \times T}^{l}}$ at the \textit{l}-th pyramid layer as an example, we first employ a 1D transposed convolution to upsample the feature $P^{l + 1}$ from the \textit{l}+1-th pyramid layer by a factor of 2 and leverage a 1D convolution to refine $P^{l}$ to obtain features $f_{l+1}$ and $f_{l}$.   

Since $f_{l + 1}$ may contain some irrelative information for classification, an attention weight $W_{l + 1}$ is learned from the element-wise summation of $f_{l}$ and $f_{l+1}$ as
\begin{equation}
	W_{l + 1} = Sigmoid\left( FC\left( GAP\left( ReLU\left( f_{l + 1} + f_{l} \right) \right) \right) \right)
\end{equation}
where $ReLU$, $GAP$, $FC$ and $Sigmoid$ denote the ReLU activation function, global average pooling, fully connected layer and sigmoid function, respectively. 

Then, the important and useful features from $f_{l+1}$ are selected by $W_{l+1}$ and combined with $P^{l}$ as
\begin{equation}
	f_{cls}^{l} = P_{l} + f_{l + 1} \otimes W_{l + 1}
\end{equation}

At last, the feature $f_{cls}^{l}$ is fed into a lightweight network composed of three 1D convolutions, layer normalization (LN) and ReLU activation ( for the first 2 blocks ). A sigmoid function is applied to yield the coarse classification result ${\hat{a}}^{l} \in R^{{C \times T}^{l}}$, where ${\hat{a}}_{t}^{l} \in R^{C}~\left( t = 1,\ldots,T^{l} \right)$ represents the probabilities of the \textit{t}-th feature vector belongs to the \textit{C} action categories in $P_{l}$. 

\textbf{Decoupled Regression Module.} Different from the classification task, localization requires the feature with temporal details to regress the start and end times of an action. Nevertheless, most existing TAL methods rely solely on single scale feature from each pyramid layer for localization, which may lose some important information. 

To address this issue, our Decoupled Regression Module (DRM) leverages the temporally rich feature from lower pyramid layer, which provides more detailed and boundary-specific information, to help action localization. As depicted in Figure 4, the DRM at the \textit{l}-th pyramid layer first employs a 1D convolution to downsample the feature from the \textit{l}-1-th layer by a factor of 2. Simultaneously, a 1D convolution is also utilized to refine the feature $P_{l}$. As a result, we can get two features $f_{l - 1}$ and $f_{l}$.

Next, similar to the DCM, the features $f_{l - 1}$ and $f_{l}$ are added to learn an attention weight $W_{l - 1}$ through a series of operations as 
\begin{equation}
	W_{l - 1} = Sigmoid\left( FC\left( GAP\left( ReLU\left( f_{l - 1} + f_{l} \right) \right) \right) \right)
\end{equation}
where the meanings of $ReLU$, $GAP$, $FC$ and $Sigmoid$ are the same as those in Eq. 5.

Based on $W_{l-1}$, the crucial features for boundary regression in  $f_{l-1}$ are selected and combined with $P^l$ as
\begin{equation}
	f_{reg}^{l} = P_{l} + f_{l - 1} \otimes W_{l - 1}
\end{equation}

Our DRM employs the Trident-head \cite{Tridet} with relative information to get the coarse boundary regression results from feature $f_{reg}^{l}$.
The boundary predictions ${\hat{s}}^{l} \in R^{T^{l}}$ and ${{\hat{e}}^{l} \in R}^{T^{l}}$ represent the start and end times for each time step in $P_{l}$, respectively.

\textbf{RefineHead.} While DCM and DRM leverage cross-layer information to generate distinct features for classification and localization tasks, their independent processing may lead to inconsistent predictions. Thus, we propose to refine the outputs of DCM and DRM using the cross layer features. For refining the prediction results of the \textit{l}-th pyramid layer, the features from the \textit{l}+1-th and \textit{l}-1-th layers are employed. 

We first combine the features from three pyramid layers by a simple Cross-Layer Feature Fusion (CLFF) module as  
\begin{equation}
	f_{c}^{l} = Conv\left( P^{l - 1} \right) + P^{l} + ConvTransposed\left( P^{l + 1} \right)
\end{equation}
where  $ConvTransposed( \cdot )$ and $Conv( \cdot )$ denote the 1D transpose convolution and 1D depth-wise convolution to upsample and downsample $P^{l + 1}$ and $P^{l - 1}$, respectively. 

Then, the fused $f_{c}^{l}$, which integrates the features benefit for both classification and localization, is fed into the RefineHead to explicitly refine the coarse predictions of DCM and DRM. For classification, a vector $R_{co}^{l} \in R^{T^{l} \times C}$ is obtained from $f_{c}^{l}$ by a simple network with three convolutional blocks to adjust the prediction of DCM. The refined result can be denoted by
\begin{equation}
	{\hat{a}}^{rl} = \sqrt{{\hat{a}}^{l} \times R_{co}^{l}}
\end{equation}

Meanwhile, two offsets $R_{so}^{l} \in R^{T^{l}}$ and $R_{eo}^{l} \in R^{T^{l}}$ are also learned from $f_{c}^{l}$ by a convolutional network to calibrate the regression results of DRM. Specifically, the \textit{t}-th element in $R_{so}^{l}$ is the offset between time step $t$ and the actual start boundary, given that time step \textit{t} is predicted as the start of an action. Through combining the offset $R_{so}^{l}$ with the prediction of ${\hat{s}}^{l}$, the refined start boundary can be obtained by
\begin{equation}
	{\hat{s}}^{rl} = {\hat{s}}^{l} + R_{so}^{l}
\end{equation}

Similarly, we can get the refined end boundary as
\begin{equation}
	{\hat{e}}^{rl} = {\hat{e}}^{l} + R_{eo}^{l}
\end{equation}

\subsection{Training and Inference}

In this study, the varifocal loss~\cite{zhang2021varifocalnet} and IOU loss~\cite{IoULoss} are employed to supervise the refined classification and regression outputs respectively. The loss function is defined as follows:
\begin{scriptsize} 
	\begin{equation}
		\mathcal{L} = \frac{1}{N_{pos}}{\sum\limits_{l,t}{1_{\{{c_{t}^{l} > 0}\}}\left( {\sigma_{IoU}\mathcal{L}_{VFL} + \mathcal{L}_{IoU}} \right)}} + \frac{1}{N_{neg}}{\sum\limits_{l,t}{1_{\{{c_{t}^{l} = 0}\}}\mathcal{L}_{VFL}}}
	\end{equation}
\end{scriptsize}where $\sigma_{IoU}$ is the temporal IoU between the predicted action and ground truth. $\mathcal{L}_{VFL}$ is varifocal loss to handle imbalanced samples in different action categories. $\mathcal{L}_{Iou}$ denotes IoU based regression loss and is applied only when the function $l_{\{ c_{t}^{l} > 0\}}$ indicates that the current time step \textit{t} is a positive sample within an action. $N_{pos}$ and $N_{neg}$ are the numbers of positive and negative samples, respectively. Following other studies~\cite{Tridet}, we utilize the center sampling to identify the positive samples.

At inference time, we first feed the full video sequences into our model to obtain the outputs for each instant \textit{t} across all pyramid layers. Then, the instants with classification scores exceeding a predetermined threshold \textit{$\lambda$} are processed by Soft-NMS~\cite{Soft-NMS} to yield the final output.

\section{Experiments}

We conduct experiments on five challenging datasets including THUMOS14~\cite{THUMOS14}, MultiTHUMOS~\cite{MultiTHUMOS}, EPIC-KITCHEN-100~\cite{EPIC-Kitchens100} ActivityNet-1.3~\cite{ActivityNet-1.3}, and HACS~\cite{HACS} to validate the effectiveness of our method. 

\subsection{Evaluation Metric}
To evaluate the performance of our CLTDR-GMG, we employ the widely adopted  mAP metric across various temporal IOU (tIoU) thresholds. In line with the prevalent methodologies, the thresholds for THUMOS14, MultiTHUMOS, and EPIC-KITCHENS-100 datasets are set as [0.3:0.1:0.7], [0.1:0.1:0.9], and [0.1:0.1:0.5], respectively. For ActivityNet-1.3 and HACS, we report the results at tIoU thresholds [0.5,0.75,0.95] and the average mAP obtained from thresholds [0.5:0.05:0.95].

\subsection{Implementation Details}
We employ the AdamW with warm-up and a cosine annealing learning rate schedule for model optimization. The number of feature pyramid layers is set to $L$=6 for THUMOS14, MultiTHUMOS and EPIC-KITCHENS-100, and $L$=7 for ActivityNet-1.3 and HACS. Given the absence of higher and lower layers for the top and bottom layers in feature pyramid, our CLTDR module is applied exclusively to intermediate feature layers with shared parameters. We conduct our experiments using Python 3.8, PyTorch 2.0, and CUDA 11.8 on a NVIDIA RTX 4090 GPU. \textit{More details about the model implementation can be found in supplementary material.} 

\subsection{Results and Analysis}
\textbf{THUMOS14.}
Following most approaches~\cite{ActionFormer,Tridet}, we leverage the pre-trained two-stream I3D~\cite{I3D} on Kinetics~\cite{Kinetics} to extract features from the THUMOS14 dataset. To ensure a fair comparison with recent state-of-the-art methods, we also employ VideoMAEv2~\cite{VideoMAEv2} and InterVideo2~\cite{wang2024internvideo2}, which are extensively pre-trained large models, for feature extraction. The performance of various TAL methods is shown in Table 1. It is evident that recent methods based on innovative techniques (such as Actionformer, TriDet and ActionMamba) outperform other counterparts. Moreover, the performance of our proposed CLTDR-GMG is superior to all existing methods. Specifically, our method achieves average mAPs of 69.9\%,  71.8\%, and 74.3\% with I3D, VideoMAE V2, and InterVideo2 features, which are at least 0.6\% , 1.7\%, and 1.3\% higher than those of previous SOTA approaches. 
\begin{table}[htbp]
	\scriptsize
	\centering
	\label{tab:table1}
	\tabcolsep=0.07cm
	\begin{tabular}{c|cccccc}
		\hline
		Method                    & 0.3  & 0.4  & 0.5  & 0.6  & 0.7  & Avg. \\ \hline
		TALLFormer~\cite{Tallformer} $\ddagger$          & 76.0 & —    & 63.2 & —    & 34.5 & 59.2 \\
		ActionFormer~\cite{ActionFormer}       & 82.1 & 77.8 & 71.0 & 59.4 & 43.9 & 66.8 \\
		TemporalMaxer~\cite{TemporalMaxer}    & 82.8 & 78.9 & 71.8 & 60.5 & 44.7 & 67.7 \\
		TFFormer~\cite{TFFormer}   & 82.1 & 78.9 & 72.0 & 60.8 & 44.9 & 67.8 \\
		TransGMC~\cite{TransGMC}     & 82.3 & 78.8 & 71.4 & 60.0 & 45.1 & 67.5 \\
		TriDet~\cite{Tridet}             & 83.6 & 80.1 & 72.9 & 62.4 & 47.4 & 69.3 \\
		\textbf{CLTDR-GMG}                     & \textbf{84.1} & \textbf{80.3} & \textbf{73.6} & \textbf{62.4} & \textbf{48.2} & \textbf{69.9} \\
		ActionFormer~\cite{ActionFormer} $\dagger$        & 84.0 & 79.6 & 73.0 & 63.5 & 47.7 & 69.6 \\
		TriDet~\cite{Tridet} $\dagger$         & 84.8 & 80.0 & 73.3 & 63.8 & 48.8 & 70.1 \\
		\textbf{CLTDR-GMG} $\dagger$                     & \textbf{85.7} & \textbf{81.3} & \textbf{75.5} & \textbf{65.3} & \textbf{51.0} & \textbf{71.8}\\
		ActionFormer~\cite{ActionFormer} $\S$       & 82.3 & 81.9 & 75.1 & 65.8 & 50.3 & 71.9 \\
		TemporalMaxer~\cite{TemporalMaxer}  $\S$ $\mp$ & 87.4 & 83.1 & 76.6 & 65.7 & 49.6 & 72.5 \\
		ActionMamba~\cite{actionmamba}  $\S$ & 86.9 & 83.1 & 76.9 & 65.9 & 50.8 & 72.7 \\
		TriDet~\cite{Tridet}  $\S$ $\mp$ & 86.9 & 83.4 & 76.6 & 66.3 & 51.7 & 73.0 \\
		\textbf{CLTDR-GMG} $\S$                     & \textbf{87.0} & \textbf{84.0} & \textbf{78.4} & \textbf{67.9} & \textbf{54.0} & \textbf{74.3} \\ \hline
	\end{tabular}
	\caption{Performance comparison on THUMOS14 dataset. $\ast $: TSN features. $\ddagger$:Swin Transformer features. $\dagger$: VideoMAEv2 features. $\S$: InterVideo2-6B features. Others: I3D features. $\mp$:indicates our implementation.}
\end{table}

\begin{table}[htbp]
	\scriptsize
	\centering
	\label{tab:table2}
	\begin{tabular}{c|cccc}
		\hline
		Method                					& 0.2  & 0.5  & 0.7  & Avg. \\ \hline
		PointTAD~\cite{PointTAD} 			& 39.7 & 24.9 & 12.0 & 23.5 \\
		ActionFormer~\cite{ActionFormer}       & 46.4 & 32.4 & 15.0 & 28.6 \\
		TemporalMaxer~\cite{TemporalMaxer} 		& 47.5 & 33.4 & 17.4 & 29.9 \\
		TriDet~\cite{Tridet}      				& 49.1 & 34.3 & 17.8 & 30.7 \\
		\textbf{CLTDR-GMG}                					& \textbf{56.7} & \textbf{42.2} & \textbf{24.1} & \textbf{37.1} \\
		TriDet~\cite{Tridet}$\ast$               					& 55.7& 41.0 & 23.5 & 36.2 \\
		\textbf{CLTDR-GMG}$\ast$               					& \textbf{62.5} & \textbf{49.1} & \textbf{30.2} & \textbf{42.7} \\
		TriDet~\cite{Tridet}$\dagger$      				& 57.7 & 42.7 & 24.3 & 37.5 \\
		\textbf{CLTDR-GMG}$\dagger$                					& \textbf{64.9} & \textbf{51.4} & \textbf{32.9} & \textbf{44.9} \\ \hline
	\end{tabular}
	\caption{Performance comparison on MultiTHUMOS dataset. $\ast$: I3D (RGB+Flow) features. $\dagger$: VideoMAEv2 features. Others: I3D (only RGB) features. }
\end{table}
\textbf{MultiTHUMOS.} Similar to the experiment on THUMOS14, we employ I3D and VideoMAEv2 to extract video features on this dataset. The results in Table 2 clearly show that our CLTDR-GMG significantly outperforms other methods. Additionally, we find that the more sophisticated VideoMAEv2 features can boost the TAL performance of our CLTDR-GMG and some other approaches, which is consistent with the results in Table 1. For instance, the VideoMAEv2 features lead to a 2.2\% improvement in average mAP for our method.

\textbf{EPIC-KITCHENS-100.} Table 3 shows the performance obtained by different TAL methods using the features extracted by pre-trained SlowFast network~\cite{SlowFast}. It can be seen that our CLTDR-GMG achieves average mAPs of 26.0\% and 24.5\% for the verb and noun subtasks, which are both superior to previous methods on these challenging datasets.
\begin{table}[htbp]
	\scriptsize
	\centering
	\setlength{\tabcolsep}{1mm}
	\label{tab:table3}
	\tabcolsep=0.06cm
	\begin{tabular}{c|c|cccccc}
		\hline
		Task                & Method                                                   & 0.1  & 0.2  & 0.3  & 0.4  & 0.5  & Avg. \\ \hline
		\multirow{6}{*}{\textit{V.}} 
		& ActionFormer~\cite{ActionFormer}     & 26.6 & 25.4 & 24.2 & 22.3 & 19.1 & 23.5 \\
		& TemporalMaxer~\cite{TemporalMaxer}   & 27.8 & 26.6 & 25.3 & 23.1 & 19.9 & 24.5 \\
		& TriDet~\cite{Tridet}              & 28.6 & 27.4 & 26.1 & 24.2 & 20.8 & 25.4 \\
		& TFFormer~\cite{TFFormer}                & 28.8 & 27.7 & 26.1 & 24.7 & 20.5 & 25.6 \\
		& TransGMC~\cite{TransGMC}                & 27.8 & 26.7 & 25.5 & 23.6 & 20.7 & 24.9 \\
		& \textbf{CLTDR-GMG}                                                   & \textbf{29.5} & \textbf{28.6} & \textbf{27.0} & \textbf{24.3} & \textbf{20.7} & \textbf{26.0} \\ \hline
		\multirow{6}{*}{\textit{N.}} 
		& ActionFormer~\cite{ActionFormer}    & 25.2 & 24.1 & 22.7 & 20.5 & 17.0 & 21.9 \\
		& TemporalMaxer~\cite{TemporalMaxer}   & 26.3 & 25.2 & 23.5 & 21.3 & 17.6 & 22.8 \\
		& TriDet~\cite{Tridet}             & 27.4 & 26.3 & 24.6 & 22.2 & 18.3 & 23.8 \\
		& TFFormer~\cite{TFFormer}               & 27.2 & 25.9 & 24.2 & 21.7 & 17.9 & 23.4 \\
		& TransGMC~\cite{TransGMC}              & 26.4 & 25.2 & 23.4 & 21.4 & 18.1 & 22.9 \\
		& \textbf{CLTDR-GMG}                                                    & \textbf{28.2} & \textbf{26.9} & \textbf{25.2} & \textbf{22.7} & \textbf{19.4} & \textbf{24.5} \\ \hline
	\end{tabular}
	\caption{ Performance comparison on EPIC-KITCHENS-100 dataset. \textit{V.} and \textit{N.} denote the verb and noun sub-tasks.}
\end{table}

\textbf{ActivityNet-1.3.} For the experiment on ActivityNet-1.3, we use TSP R(2+1)D~\cite{TSP} as the pre-trained model to extract video features, which is consistent with the methodology used in several recent studies~\cite{Tridet}. As shown in Table 4, our method outperforms other methods.
\begin{table}[htbp]
	\centering
	\scriptsize
	\tabcolsep=0.12cm
	\label{tab:table4}
	\begin{tabular}{c|cccc}
		\hline
		\multirow{2}{*}{Method}              & \multicolumn{3}{c}{tIoU} & Avg.          \\ \cline{2-5} 
		& 0.5    & 0.75   & 0.95   & 0.5:0.05:0.95 \\ \hline
		ReAct~\cite{ReAct}$\ast$             & 49.6   & 33.0     & 8.6    & 32.6          \\
		TadTR~\cite{TadTR}$\ast$            & 51.3   & 35.0     & 9.5    & 34.6          \\
		TadTR~\cite{TadTR}$\dagger$             & 53.6   & 37.5   & 10.5   & 36.8          \\
		TALLFormer~\cite{Tallformer}$\ddagger$        & 54.1   & 36.2   & 7.9    & 35.6          \\
		TFFormer~\cite{TFFormer}     & 54.4   & 36.7   & 7.5    & 35.8          \\
		ActionFormer~\cite{ActionFormer}$\dagger$      & 54.7   & 37.8   & 8.4    & 36.6          \\
		TransGMC~\cite{TransGMC}$\dagger$    & 54.8   & 37.6   & 8.5    & 36.7          \\
		TriDet~\cite{Tridet}$\dagger$            & 54.7   & 38.0     & 8.4    & 36.8          \\
		\textbf{CLTDR-GMG}$\dagger$                            & \textbf{55.0}   & \textbf{38.0}  & \textbf{8.6}    & \textbf{37.1}          \\ \hline
	\end{tabular}
	\caption{Performance comparison on ActivityNet1.3 dataset. $\ast$: TSN features. $\ddagger$: Swin Transformer features. $\dagger$: R(2+1)D features. Others: I3D features. }
\end{table}

\textbf{HACS.}
On HACS dataset, we use the I3D~\cite{I3D} features from RGB stream and the SlowFast~\cite{SlowFast} features from TCANet~\cite{TCANet} in our experiments. As shown in Table 5, our method achieves average mAPs of 37.2\% with I3D features and 39.2\% with SlowFast features, which outperforms the most recent TriDet model. 
\begin{table}[htbp]
	\centering
	\scriptsize
	\label{tab:table5}
	\begin{tabular}{c|cccc}
		\hline
		Method           & 0.5  & 0.75 & 0.95 & Avg.  \\ \hline
		TadTR~\cite{TadTR}      & 47.1 & 32.1 & 10.9 & 32.1 \\
		TALLFormer~\cite{Tallformer}$\ddagger$  & 55.0   & 36.1 & 11.8 & 36.5 \\
		TCANet~\cite{TCANet}$\dagger$     & 54.1 & 37.2 & 11.3 & 36.8 \\
		TriDet~\cite{Tridet}     & 54.5 & 36.8 & 11.5 & 36.8 \\
		\textbf{CLTDR-GMG}        & \textbf{55.2} & \textbf{37.3} & \textbf{11.8} & \textbf{37.2} \\
		TriDet~\cite{Tridet}$\dagger$     & 56.7 & 39.3 & 11.7 & 38.6 \\
		\textbf{CLTDR-GMG}$\dagger$       & \textbf{57.6} & \textbf{39.9} & \textbf{12.0} & \textbf{39.3} \\ \hline
	\end{tabular}
	\caption{Performance comparison on HACS dataset. $\ddagger$: Swin Transformer features.
		$\dagger$: SlowFast features. Others: I3D features.}
\end{table}
\subsection{Ablation Study}
We perform various ablation studies on THUMOS14 dataset using InterVideo2-6B features to assess the effectiveness of each component in our CLTDR-GMG.

\textbf{Ablation on GMG.}
First, we conduct an ablation experiment to demonstrate the effectiveness of each temporal granularity branch in our GMG module. Table 6 shows that our method does not achieve satisfactory performance when using information from instant, local or global granularity individually. Furthermore, combining global granularity information with either instant or local granularity information yields better results than combining the information of instant and local granularities. This substantiates that global granularity information is crucial for the TAL task. Finally, our method achieves optimal performance when integrating information from all three temporal granularities.
\begin{table}[!h]
	\scriptsize
	\centering
	\begin{tabular}{c|ccccccc}
		\hline
		Instant   & \Checkmark   & \ding{55}  & \ding{55}  & \Checkmark & \Checkmark & \ding{55}  & \Checkmark \\  
		Local   & \ding{55} & \Checkmark & \ding{55} & \Checkmark & \ding{55} & \Checkmark  & \Checkmark \\ 
		Global & \ding{55}  & \ding{55}  & \Checkmark  & \ding{55}  & \Checkmark & \Checkmark & \Checkmark \\ 
		\hline
		Avg. &73.4 &73.4 &73.5 &73.6 & 73.8 &73.9 &74.3 \\ 
		\hline
	\end{tabular}
	\caption{The analysis of different temporal granularities.}
\end{table}

\begin{table}[h]
	\scriptsize
	\centering
	\begin{tabular}{ccccc}
		\hline
		Method       & 0.3  & 0.5  & 0.7  & Avg. \\ \hline
		without Gate & 87.0 & 77.7 & 53.4 & 73.9 \\
		with Gate    & 87.3 & 77.9 & 53.9 & 74.3 \\ \hline
	\end{tabular}
	\caption{The impact of gated mechanism in GMG.}
\end{table}

\begin{table}[!h]
	\scriptsize
	\centering
	\begin{tabular}{cccccc}
		\hline
		Global-Level   & 0.3  & 0.5  & 0.7  & Avg. & \# Params \\ \hline
		FFT            & 87.3 & 77.9 & 53.9 & 74.3 & 9.5M          \\
		Self-Attention & 87.5 & 78.0   &53.5 & 74.2 & 26.1M         \\ \hline
	\end{tabular}
	\caption{The comparison of global feature extractor in GMG.}
\end{table}
We then explore the impact of gated mechanism in GMG module. As shown in Table 7,  our method achieves a performance improvement of 0.4\% when the gated mechanism is used. This confirms the effectiveness of gated mechanism for adaptive feature selection and redundancy elimination.

Finally, we replace the FFT-based global feature extractor in our GMG with vanilla self-attention used by Transformer. Table 8 reveals that while vanilla self-attention achieves similar performance as our method, it incurs a significantly higher parameter count, approximately 2.7 times that of our FFT-based approach. 

\textbf{Ablation on CLTDR.} To investigate the impact of feature pyramid layers on classification and regression decoupling, we conducted ablation studies as shown in Table 9. Our results indicate that fusing features from higher pyramid layer enhances classification performance due to the richer semantic information in them. For regression tasks, incorporating features from lower layer contributes to more performance gains, as they provide more detailed information. We further find that incorporating both higher and lower pyramid layers leads to a slight performance degradation, which suggests that excessive information may be redundant or even detrimental for task decoupling.

Then, the effectiveness of RefineHead in CLTDR is studied. The results in Table 10 indicates that the performance of our method can be improved by refining the predictions of classification and localization tasks.

At last, the attention based cross layer information fusion strategy in DCM and DRM is compared with addition and concatenation. From Table 11, it is clear that the attention mechanism achieves better performance than other two information fusion manners. 

\textit{Due to page limitations, additional experimental results and analyses of our method are provided in the supplementary material accompanying this paper.}
\begin{table}[t]
	\scriptsize
	\centering
	\begin{tabular}{cccc|cccc}
		\hline
		\multicolumn{4}{c|}{classification} & \multicolumn{4}{c}{regression} \\
		$P_{l+1}$    & $P_{l}$   & $P_{l-1}$   & mAP    & $P_{l+1}$  & $P_{l}$  & $P_{l-1}$  & mAP   \\ \hline
		& \Checkmark     &         & 73.3   &        & \Checkmark    &        & 73.4  \\
		\Checkmark        & \Checkmark    &         & 74.3   &        & \Checkmark    & \Checkmark      & 74.3  \\
		& \Checkmark     & \Checkmark       & 73.8   & \Checkmark      & \Checkmark    &        & 73.6  \\
		\Checkmark       & \Checkmark     & \Checkmark       & 74.0   & \Checkmark      & \Checkmark    & \Checkmark      & 73.2  \\ \hline
	\end{tabular}
	\caption{Ablation of pyramid layers in CLTDR.}
\end{table}
\begin{table}[t]
	\scriptsize
	\centering
	\begin{tabular}{lllll}
		\hline
		\multicolumn{1}{c}{Method}        & 0.3  & 0.5  & 0.7  & Avg. \\ \hline
		without refinement                & 87.1 & 77.3 & 53.1 & 73.5 \\
		only refine classification    & 87.4 & 77.8 & 53.9 & 73.9 \\
		only refine regression  & 87.6 & 77.3 & 52.4 & 73.8 \\
		refine classification and regression & 87.3 & 77.9 & 53.9 & 74.3 \\ \hline
	\end{tabular}
	\caption{The effectiveness of refinement in CLTDR.}
\end{table}
\begin{table}[!h]
	\scriptsize
	\centering
	\begin{tabular}{ccccc}
		\hline
		Method      & 0.3  & 0.5  & 0.7  & Avg. \\ \hline
		Concatenation & 86.7 & 76.7 & 53.3 & 73.4 \\
		Addition         & 86.8 & 77.3 & 52.8 & 73.5 \\
		Attention   & 87.3 & 77.9 & 53.9 & 74.3 \\ \hline
	\end{tabular}
	\caption{The comparison of fusion strategies in CLTDR.}
\end{table}
\section{Conclusion}
In this paper, we proposed a GMG based encoder and a CLTDR based decoder for TAL problem. Experiments conducted on five datasets demonstrate that our method outperforms the previous approaches and achieves state-of-the-art performance. Moreover, several ablation studies are also carried out to validate the components in our method. 
\section{Acknowledgments}
This work is supported by the NSFC under Grant 62272096 and the Jilin Provincial Science and Technology Department under Grant 20230201079GX.
\small\bibliography{aaai25}

\appendix
\normalsize
\onecolumn

\section{Supplementary Material}
\subsection{A. Comparison with Other Studies}
The Fourier transform has been adopted for global feature extraction in some studies~\cite{FNet,GFNet,AFNO}. However, our GMG module differs from them in the following aspects. First, FNet~\cite{FNet} simply adds the input feature and their Fourier spectrum without filter learning or inverse Fourier transform. Thus, it is less flexible than our method in which a global filter can be learned to refine the feature in frequency domain. Second, although GFNet~\cite{GFNet} directly learns global filters to capture the feature relation in Fourier domain, it can only be applied to the tasks with a fixed number of vision tokens. This makes it unsuitable for TAL problem in which the lengths of various videos are different. Third, the filters learned in AFNO~\cite{AFNO} are invariant to the number of vision tokens. Nevertheless, AFNO is designed for image based vision tasks, such as image classification and segmentation, rather than video feature sequences based TAL task. Finally, none of the above methods explicitly takes the instant and local information into consideration.

The task decoupling are widely used in one stage object detectors~\cite{RCNN,yolox}. However, as we have analyzed, most existing methods only disentangle the classification and localization tasks at parameter level. Although TSCODE~\cite{tscode} leverage the features from multiple pyramid layers for task decoupling, there still exist two key differences between it and our CLTDR. On the one hand, TSCODE utilizes simple addition and concatenation to fuse the features of different layers, which potentially induce some irrelative information for classification and localization tasks. In contrast, our CLTDR employs an attention mechanism to selectively integrate crucial features from different pyramid layers to facilitate the task decoupling. On the other hand, TSCODE only considers the task decoupling problem, whereas our CLTDR takes both the task decoupling and refinement into account. To the best of our knowledge, the proposed CLTDR is the first work to simultaneously address the task decoupling and refinement by cross layer features in TAL domain.

\subsection{B. Details of Datasets}
\textbf{THUMOS14} is a dataset for evaluating video action detection and recognition algorithms. It consists of multiple video clips, each containing a scene in which a person or object performs a specific action. The Thumos14 dataset for temporal action localization and detection contains unsegmented videos of 20 action categories and annotation information for the temporal behavioral segments, The action categories in the dataset include running, jumping, skating, walking, cycling, swimming, and so on.

\textbf{MultiTHUMOS} is an extension of the THUMOS dataset. The dataset includes 38,690 annotations for 65 action classes. The annotations are applied to 400 videos in THUMOS'14 motion detection dataset, with an average of 1.5 labels per frame and 10.5 action classes per video.
EPIC-KITCHENS-100 is a large-scale first-view video dataset for computer vision that contains more than 100 hours of videos on daily kitchen activities. Examples of actions include cooking, cleaning, and food preparation. Totally, the dataset contains over 90,000 action clips, 97 verb categories and 300 noun categories.

\textbf{ActivityNet-1.3} dataset is a large-scale temporal action localization and detection benchmark, which includes 200 different types of actions and a total of 849 hours of YouTube videos. The dataset consists of 19,994 untrimmed videos which are divided into three subsets for training, validation, and testing in a 2:1:1 ratio. Each action category includes an average of 137 untrimmed videos, and each video has an average of 1.41 actions with temporal boundaries annotated. The action categories are primarily divided into seven major classes: personal care, diet, home activities, care and assistance, work, social entertainment, and sports and exercise. ActivityNet is widely used in computer vision due to its large scale and high-quality annotations.

\textbf{HACS} dataset is derived from YouTube and specifically designed for the task of human action recognition. It uses a taxonomy of 200 action classes, which is identical to that of the ActivityNet-v1.3 dataset. The dataset includes an impressive 504,000 videos, each of which is no longer than 4 minutes, with an average duration of 2.6 minutes. To ensure a diverse and balanced sample, a total of 1.5 million 2-second clips were extracted using both uniform random sampling and image classifier consensus. Out of these, 0.6 million clips have been labeled as positive examples, while 0.9 million serve as negative samples.

\subsection{C. Implementation Details}
In our model, we follow Ref.~\cite{Tridet}to set the parameter B (i.e., the number of bins) in Trident-head as 16, 16, 16, 12 and 14 for THUMOS14, MultiTHUMOS, EPIC-KICHENS-100, ActivityNET and HACS datasets, respectively. The kernel sizes of 1D depth-wise convolution and transposed convolution utilized in global filter learning, DCM, DRM and RefineHead are all set as 3. The detailed description of experiments conducted on each dataset is presented below.

\textbf{THUMOS14.} For I3D, VideoMAE V2, and InterVideo2-6B features on this dataset, the initial learning rate and mini-batch size are set as 0.0001 and 2. The number of training epochs is respectively set as 40, 62, and 57, with warmup periods of 20, 30, and 32 epochs. Weight decay is set to 0.035, 0.055, and 0.05 for the three features. 

\textbf{MultiTHUMOS.} For this dataset, the initial learning rate is set as 0.00015, the mini-batch size is 2, the number of training epochs is 60 (with 25 warmup epochs) and the weight decay is 0.03. 

\textbf{EPIC-KITCHENS-100.} For verb and noun sub-tasks in EPIC-KITCHENS, the initial learning rate is set as 0.0001. The number of training epochs is 28 and 25 (with a warmup of 10 and16 epochs), respectively. The weight decay is set to 0.05 and 0.04 for verb and noun sub-tasks. 

\textbf{ActivityNet-1.3.} The initial learning rate is set as 0.001 and the mini-batch size is set as 16 for ActivityNet-1.3 dataset. The number of training epochs is 15 (with 10 warmup epochs ) and the weight decay is set to 0.045. 

\textbf{HACS.} The initial learning rate, mini-batch size and weight decay are set as 0.001, 16, and 0.03, respecviely. The number of training epochs is 17 and 15 for I3D and SlowFast features, respectively. 

\subsection{D. Feature Extraction by Pre-trained Backbone}

\textbf{THUMOS14.} Following most other approaches~\cite{TemporalMaxer,ActionFormer,Tridet}, we leverage the pre-trained two-stream I3D~\cite{I3D} on Kinetics~\cite{Kinetics} to extract features from THUMOS14 dataset. To achieve feature extraction, we feed 16 consecutive frames into I3D network using a sliding window with a stride of 4. The extracted feature with 1024 dimensions is obtained from the last fully connected layer. We then concatenated the two-stream features to form a 2048-dimensional input for our model. For the feature extraction by VideoMAEv2, we also input consecutive 16 frames into VideoMAEv2 with a stride of 4 to obtain a 1408-dimensional feature vector. For feature extraction with InternVideo2-6B, we input consecutive 16-frame segments with a stride of 4. We use the 3200-dimensional feature vector from the 7th layer of InternVideo2 as the extracted features, without incorporating any additional fusion.

\textbf{MultiTHUMOS.} The specific parameter settings for feature extraction with I3D and VideoMAEv2 networks for this dataset are consistent with those used for THUMOS14.

\textbf{EPIC-KITCHENS-100.} Following previous work~\cite{TemporalMaxer,ActionFormer,Tridet}, we extract the video features using SlowFast network~\cite{SlowFast} pre-trained on EPICKitchens~\cite{EPIC-Kitchens100}. We utilize consecutive 32 video frames with a stride of 16 as input sequence to the SlowFast network, so that the features with 2304-dimensions can be obtained.

\textbf{ActivityNet-1.3.} We employ two-stream I3D~\cite{I3D} and TSP~\cite{TSP} for feature extraction, which also use consecutive 16 video frames with a stride of 16 as input sequence. Following~\cite{ActionFormer}, the extracted I3D and TSP features are downsampled into a fixed length of 160 and 192 using linear interpolation, respectively.

\textbf{HACS.} Following~\cite{Tridet}, we use the I3D~\cite{I3D} features of RGB stream and the SlowFast~\cite{SlowFast} features from TCANet~\cite{TCANet} in our experiments. For I3D feature extraction, we use each single frame as input of the I3D network. For SlowFast feature extraction, we extract a 2304-dimensional feature vector from each snippet, which contains 32 frames with a stride of 8.

\subsection{E. Additional Experimental Results}

\textbf{E.1 The effectiveness of GMG encoder}

Since our proposed Gated Multi-Granularity (GMG) can be applied as a plug-and-play encoder to other TAL models, we combine it with the decoders of ActionFormer and TriDet to demonstrate its effectiveness. In this experiment, we follow the original settings in ActionFormer and TriDet (including hyper-parameters, loss function and so on) but merely replace their encoders (i.e., Transformer and SGP) with GMG. Table 12 shows that our GMG based encoder enhances the average mAP by 0.8\% in ActionFormer and 0.4\% in TriDet. This can be attribute to its comprehensive feature extraction from instant, local and global granularities. Nevertheless, due to the decoders in ActionFormer and TriDet only decouple the classification and localization tasks at parameter level with the same input feature, the TAL results of GMG+ActionFormer-decoder and GMG+Trident-decoder are still inferior to GMG+CLTDA.
\begin{table}[htbp]
	\centering
	\setlength{\tabcolsep}{0.8mm}
	\begin{tabular}{ccccc}
		\hline
		Method                   & 0.3  & 0.5  & 0.7  & Avg.       \\
		\hline
		GMG+ActionFormer-decoder & 87.3 & 77.0 & 49.6 & 72.7(↑0.8) \\
		GMG+Trident-decoder      & 87.1 & 77.1 & 51.6 & 73.4(↑0.4) \\
		GMG+CLTDA                & 87.3 & 77.9 & 53.9 & 74.3   \\  
		\hline 
	\end{tabular}
	\caption{The analysis of GMG encoder on THUMOS14.}
\end{table}

\textbf{E.2 The effectiveness of CLTDR decoder}

To justify the effectiveness of our proposed CLTDR which leverages cross layer features for classification and localization tasks decoupling and refinement, we integrate it with the encoders of ActionFormer, TemporalMaxer, and TriDet. Similar to the previous experiment, we keep the other settings in these three methods but solely substitute their decoders with our CLTDR. By comparing the results in Table 13, we can see that our CLTDR based decoder improves the performance of different encoders by 0.6\%–0.8\% in average mAP. 
\begin{table}[htbp]
	\centering
	\setlength{\tabcolsep}{0.5mm}
	\begin{tabular}{cccccc}
		\hline
		\multicolumn{1}{c}{Method}  & 0.3  & 0.5  & 0.7  & Avg.       \\
		\hline
		ActionFormer-encoder+CLTDR  & 85.9 & 76.7 & 51.5 & 72.6(↑0.7) \\
		TemporalMaxer-encoder+CLTDR & 87.4 & 77.8 & 51.0 & 73.1(↑0.6) \\
		Trident-encoder+CLTDR       & 86.7 & 77.6 & 53.4 & 73.8(↑0.8) \\
		ActionMamba-encoder+CLTDR  & 86.3 & 77.4 & 52.8 & 73.4(↑0.7) \\
		GMG+CLTDR                   & 87.3 & 77.9 & 53.9 & 74.3  \\
		\hline    
	\end{tabular}
	\caption{The influence of different pyramid levels on our method.}
\end{table}

\textbf{E.3 The influence of feature pyramid levels}

Table 14 shows the performance of our method with different numbers of pyramid layers. It is evident that the performance of our method improves initially with the increase in the number of pyramid layers. This indicates the significance of the feature pyramid structure in our approach. However, after achieving the optimal results with 6 pyramid layers on THUMOS14, EPIC-KITCHENS-100, and MultiTHUMOS datasets, and 7 pyramid layers on ActivityNet and HACS datasets, its mAPs begin to decline.  The underlying reason for this phenomenon might be that information about short actions with very small time steps is lost in the excessively compressed features of higher pyramid layers.
\begin{table*}[htbp]
	\centering
	\scriptsize
	\tabcolsep=0.15cm
	\begin{tabular}{c|cccc|cccc|cccc|cccc|cccc}
		\hline
		\multirow{2}{*}{Levels} &
		\multicolumn{4}{c|}{Thumos14 InterVideo2-6B} &
		\multicolumn{4}{c|}{MultiTHUMOS-VideoMAEv2} &
		\multicolumn{4}{c|}{PIC-KITCHENS-100 (verb/noun)}&
		\multicolumn{4}{c|}{ActivityNet-1.3} &
		\multicolumn{4}{c}{HACS (Slowfast)} \\
		& 0.3  & 0.5  & 0.7  & Avg. & 0.2  & 0.5  & 0.7  & Avg. & 0.1  & 0.3  & 0.5  & Avg. & 0.5  & 0.75 & 0.95 & Avg. & 0.5  & 0.75 & 0.95 & Avg. \\ \hline
		1 & 78.3 & 60.5 & 24.6 & 55.6 & 49.3 & 31.2 & 15.2 & 29.4 & 26.9/24.4 & 22.7/21.4 & 16.3/14.5 & 22.3/20.5 & 21.9 & 9.2  & 0.7  & 10.7 & 24.2 & 11.8 & 0.8  & 12.7 \\
		2 & 81.6 & 67.3 & 38.4 & 63.7 & 57.5 & 40.9 & 22.8 & 36.7 & 26.1/25.8 & 23.1/22.9 & 17.1/17.2 & 22.4/22.2 & 50.7 & 33.3 & 6.8  & 32.8 & 47.9 & 30.4 & 6.1  & 30.2 \\
		3 & 84.2 & 71.3 & 45.5 & 68.3 & 62.3 & 46.5 & 26.5 & 40.7 & 28.7/26.9 & 25.1/23.9 & 18.9/17.1 & 24.6/22.9 & 55.1 & 35.4 & 7.0  & 34.5 & 51.3 & 33.1 & 6.9  & 32.7 \\
		4 & 84.5 & 74.9 & 50.3 & 71.0 & 63.5 & 48.4 & 29.6 & 42.5 & 27.8/27.1 & 25.2/23.7 & 18.7/18.3 & 24.3/23.3 & 54.3 & 36.6 & 7.6  & 35.8 & 54.3 & 35.9 & 8.6  & 35.5 \\
		5 & 86.7 & 76.7 & 52.3 & 73.0 & 64.0 & 49.9 & 31.4 & 43.7 & 27.5/27.6 & 24.8/24.2 & 20.0/18.3 & 24.4/23.6 & 55.1 & 37.5 & 7.9  & 36.5 & 57.3 & 38.6 & 10.6 & 38.4 \\
		6 & 87.3 & 77.9 & 53.9 & 74.3 & 64.9 & 51.4 & 32.9 & 44.9 & 29.5/28.2 & 27.0/25.2 & 20.7/19.4 & 26.0/24.5 & 55.1 & 37.9 & 8.3  & 36.9 & 57.8 & 39.8 & 11.4 & 39.2 \\
		7 & 87.4 & 77.8 & 51.9 & 73.6 & 64.0 & 50.2 & 32.3 & 44.0 & 28.2/27.8 & 26.1/24.4 & 20.9/18.4 & 25.3/23.8 & 55.0 & 38.0 & 8.6  & 37.1 & 57.6 & 39.9 & 12.0 & 39.3 \\
		8 & 87.1 & 77.4 & 53.0 & 73.7 & 64.8 & 50.3 & 31.9 & 44.3 & 28.0/27.1 & 25.4/24.0 & 20.2/18.5 & 24.7/23.4 & 54.9 & 37.9 & 8.5  & 36.8 & 57.5 & 39.3 & 11.2 & 38.9 \\ \hline
	\end{tabular}
	\caption{The influence of different pyramid levels on our method.}
\end{table*}

\textbf{E.4 Computational Complexity}

We have evaluated the computational efficiency of our method by measuring the number of parameters and inference time (average over five times) on THUMOS14 dataset using an input feature size of 2304*2048. Moreover, we also compared the efficiency of our method with some other one-stage TAL approaches. From the results in Table 15, we can see that our method achieves the optimal TAL performance with moderate number of parameters and inference latency.
\begin{table}[ht]
	\centering
	\begin{tabular}{ccc}
		\hline
		Method        & \# of parameters & inference time \\ \hline
		ActionFormer  & 29.3M           & 52ms           \\
		TemporalMaxer & 7.1M            & 30ms           \\
		Tridet        & 15.9M           & 45ms           \\
		ActionMamba   & 20.4M           & 60ms           \\
		CLTDR-GMG     & 17.6M           & 48ms           \\ \hline
	\end{tabular}
	\caption{The comparison of numbers of parameters and inference time among different methods.}
\end{table}

\textbf{E.5 Visual Comparison}

To visually demonstrate the advantages of our CLTDR-GMG method, we employ a cricket bowling action video from THUMOS14 dataset as an example. From the visual comparison in Figure 5, we can see that the ActionFormer and TriDet, which disentangle the classification and localization at parameter level, cannot effectively balance the two tasks. That is, they exhibit a clear bias towards localization while compromising classification performance. The maximum classification score obtained by ActionFormer is only 0.57, corresponding to a tIOU of 0.72. Though TriDet utilizes the relative boundary information to improve the localization result, its classification performance remains similar to that of ActionFormer and suboptimal. Thanks to the cross layer task decoupling and refinement, the classification performance is dramatically improved by our method to achieve more consistent and balanced classification and localization results for each frame, as more points are located near the diagonal line in Figure 5(c). Moreover, the tIOU corresponding to the frame with best classification performance is also better than those in other two approaches.
\begin{figure*}[t]
	\centering
	\includegraphics[width=1\textwidth]{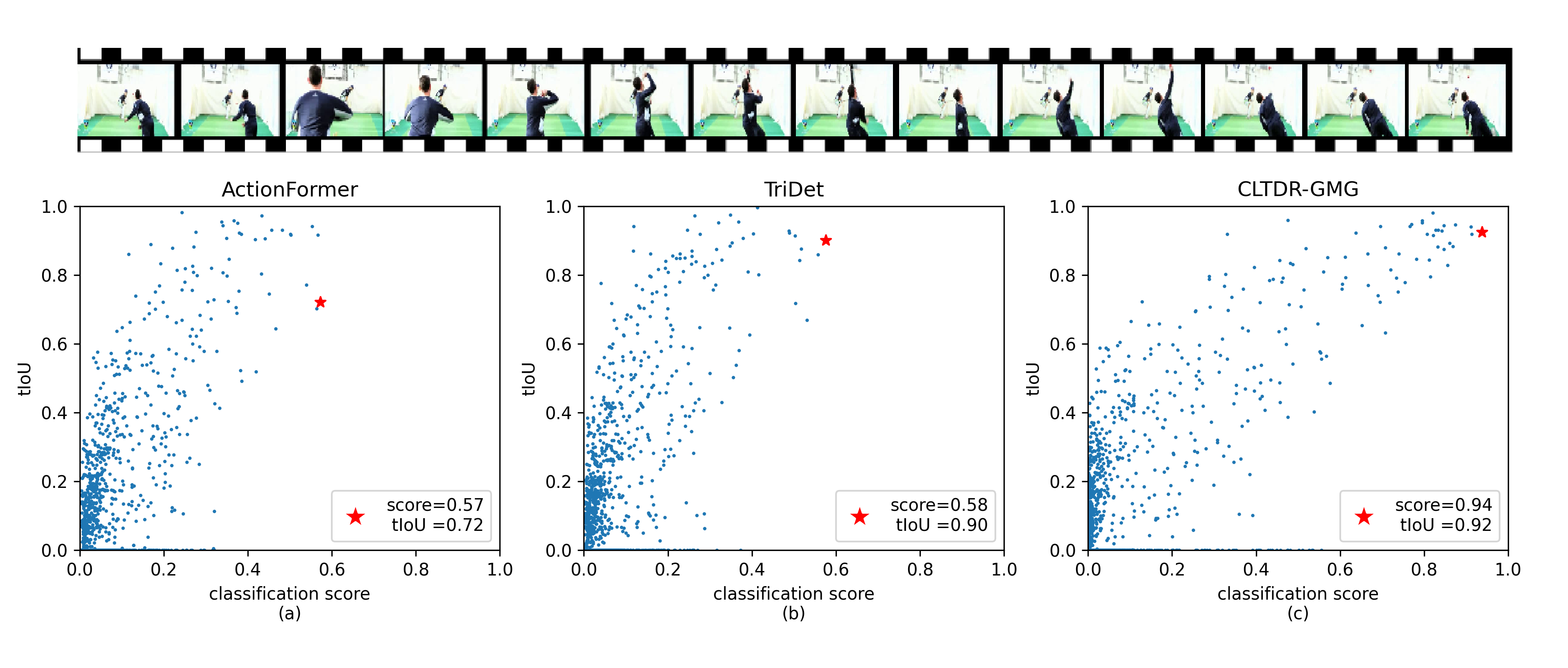} 
	\caption{The visual comparison between CLTDR-GMG and other two methods before NMS. Each point denotes the classification and localization results (i.e., classification score and tIOU) of a frame in the video. The points with the best classification scores are marked as "{$\star$}".}
\end{figure*}

\subsection{F. Error Diagnosis}
To further assess the effectiveness of our method, we conduct multiple analyses using the DETAD~\cite{DETAD} tools on the THUMOS14 dataset. The comprehensive explanations and metrics employed in DETAD's error diagnosis are described in~\cite{DETAD}.

\textbf{F.1 False Positive Analysis}

Figure 6 shows the false positive profiles and the impact of error types on the performance of our method with the proportion of different action types at various k-G values, where G represents the number of ground truths for each action category, and only the top-ranking predicted actions are retained for visualization. As depicted in Figure 6, we observe that the True Positive rate for 1G predictions exceeds 81\% at IoU=0.5. This finding suggests that our method can accurately identify and classify most actions. Furthermore, the right side of Figure 6 shows that the localization error and background confusion are two most common error types that reduce the performance of our method. This observation is consistent with results from several other studies~\cite{ActionFormer,TemporalMaxer,Tridet} . Therefore, addressing these two types of errors is critical for improving TAL performance.
\begin{figure*}[!h]
	\centering
	\includegraphics[width=0.8\linewidth]{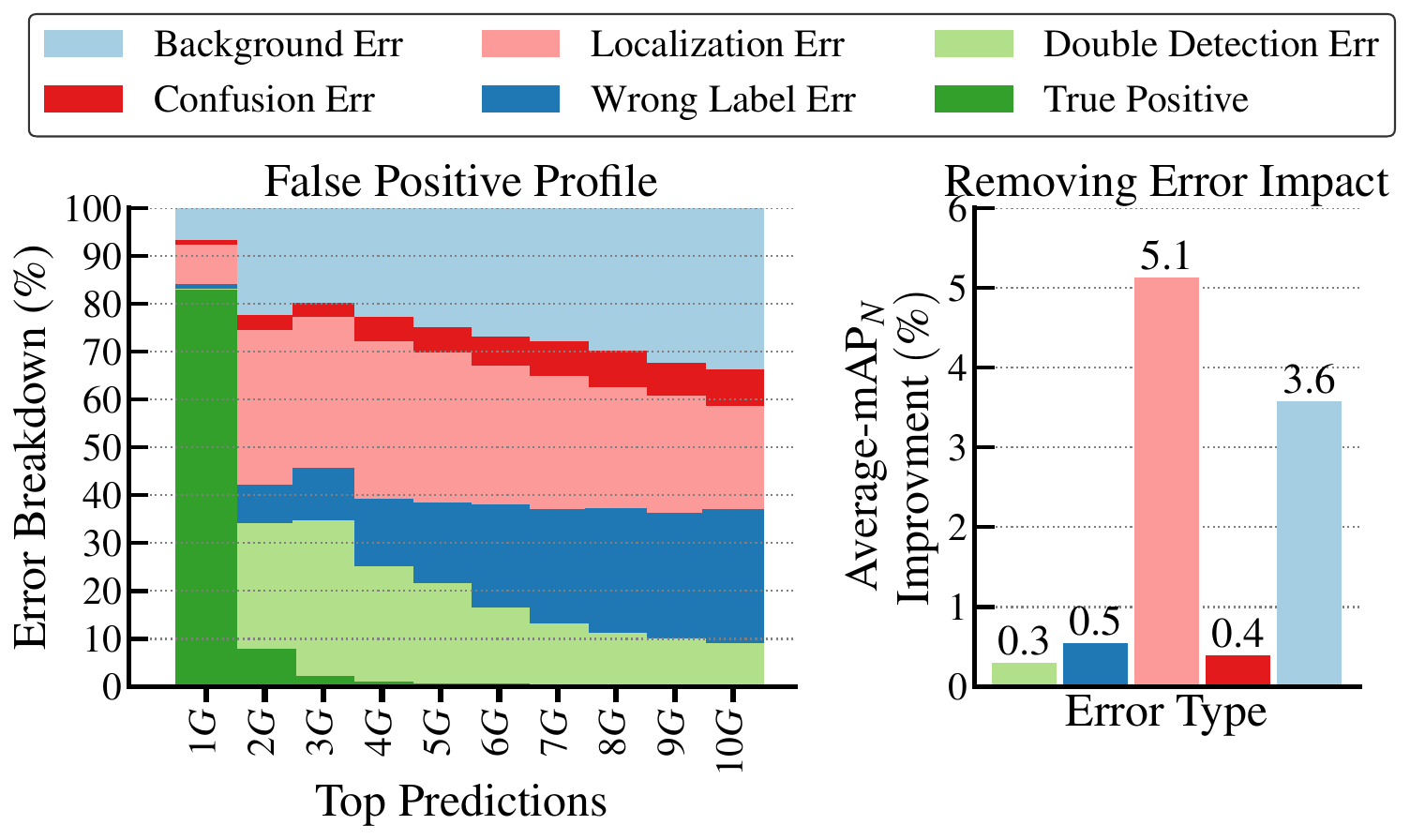} 
	\caption{The false positive profile obtained by our methods and the impact of error types.}
\end{figure*}

\begin{figure*}[!h]
	\centering
	\includegraphics[width=1\linewidth]{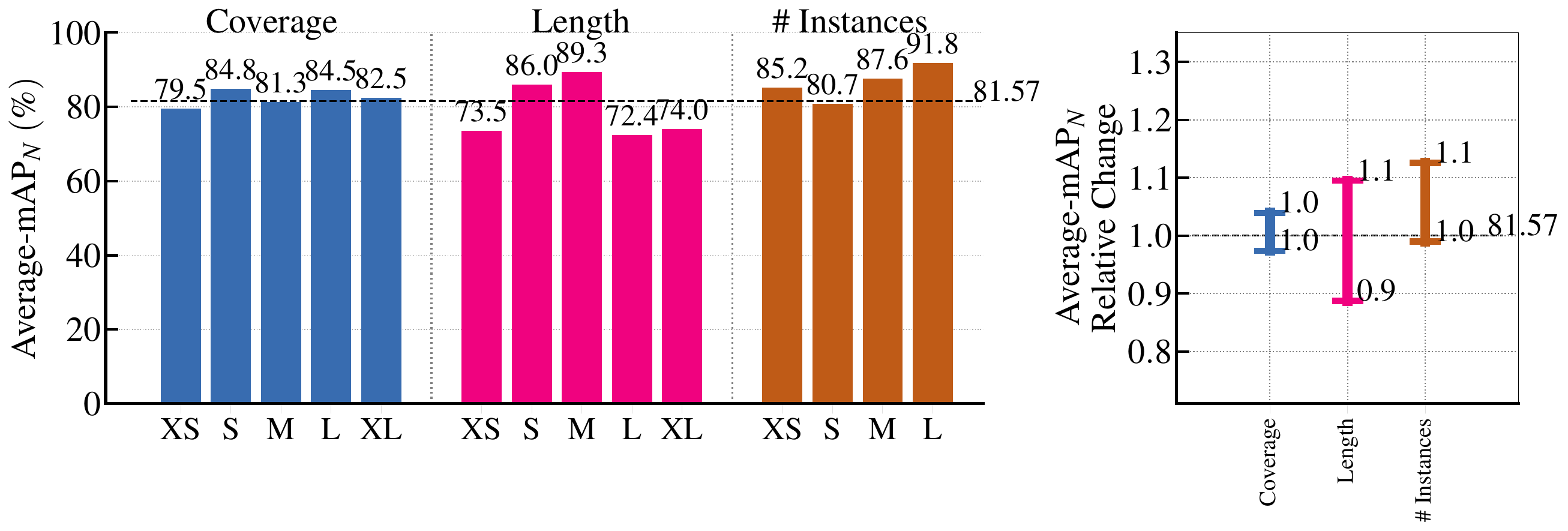} 
	\caption{The average-mAP of our method for different action metrics, where $mAP_N$ is the normalized mAP at tIoU=0.5 and the dashed line is the overall performance.}
\end{figure*}

\begin{figure*}[htbp]
	\centering
	\includegraphics[width=1\linewidth]{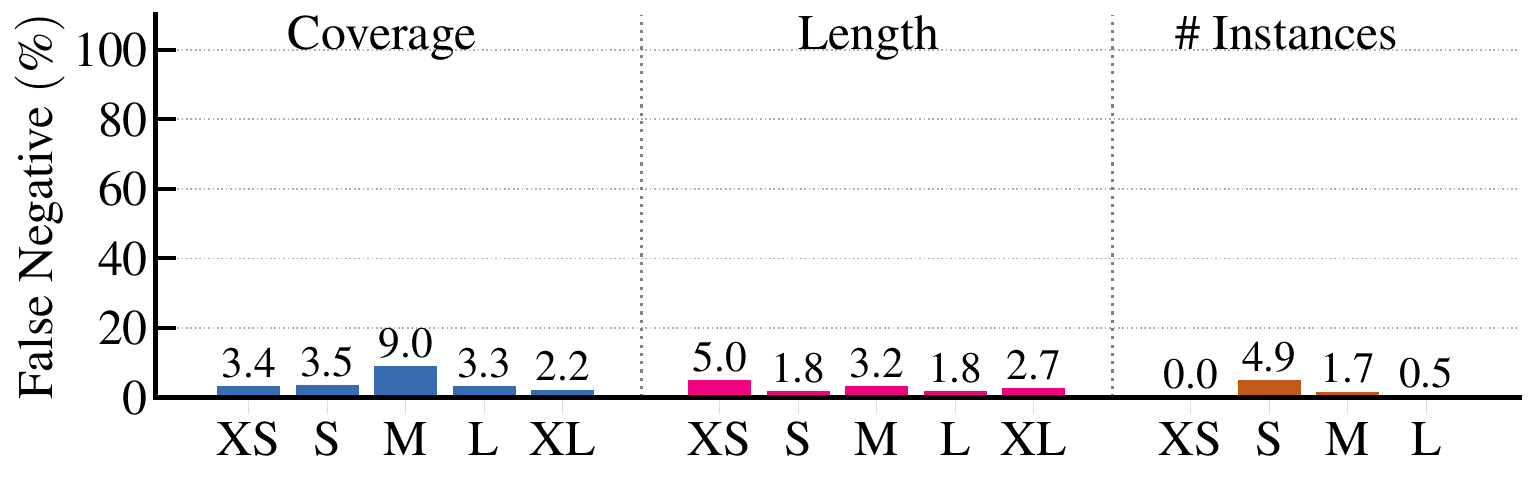} 
	\caption{The average false negative rate of our method for various characteristics of actions.}
\end{figure*}

\textbf{F.2 Sensitivity Analysis}

Figure 7 illustrates our method's sensitivity to various action characteristics. Specifically, Coverage refers to the relative length of the action (compared to the entire video containing it), Length denotes the absolute length (in seconds) of the action, and \# Instances refers to the total count of actions (from the same class) in a video. The TAL results for each metric are further divided into various categories: XS (extremely short), S (short), M (medium), L (long), and XL (extremely long). It is evident that our approach achieves satisfactory results in most action categories. 

\textbf{F.3 False Negative Analysis}

Figure 8 demonstrates that our method exhibits relatively low false negative rates for extremely long actions in Coverage. This strength can be attributed to the global information captured by the GMG encoder. Moreover, our model achieves a zero false positive rate in the XS of \# Instances metric, which demonstrates its perfect detection accuracy for the videos containing only one action.

\end{document}